\documentclass[11pt,a4paper]{article}
\usepackage[hyperref]{main}
\usepackage{times}
\usepackage{latexsym}
\usepackage{graphicx}
\usepackage{caption}
\usepackage{amsmath}
\usepackage{amssymb}
\usepackage{amsfonts}
\usepackage{booktabs}
\usepackage[bottom]{footmisc}
\usepackage{url}
\usepackage[linesnumbered,lined,algoruled]{algorithm2e}
\usepackage{multicol}
\graphicspath{{png/}}

\aclfinalcopy 

\title{Softmax Attention with Constant Cost per Token}

\author{Franz A. Heinsen \\
	{\tt franz@glassroom.com} \\ }

\date{March 28, 2024}

\DeclareMathOperator{\Attention}{Attention}
\DeclareMathOperator{\Softmax}{Softmax}
\DeclareMathOperator{\lse}{{\scriptstyle LSE}}
\DeclareMathOperator{\lcse}{{\scriptstyle LCSE}}

\newcommand{\modified}[1]{\overset{\text{modified}}{#1}}

\newcommand{\bigO}{\mathcal{O}}
\newcommand{\HofS}{H^{\scriptscriptstyle (S)}}
\newcommand{\HofZ}{H^{\scriptscriptstyle (Z)}}

\begin{document}
\maketitle

\begin{abstract}
We propose a simple modification to the conventional attention mechanism applied by Transformers: Instead of quantifying pairwise query-key similarity with scaled dot-products, we quantify it with the logarithms of scaled dot-products of exponentials. Our modification linearizes attention with exponential kernel feature maps, whose corresponding feature function is infinite dimensional. We show that our modification is expressible as a composition of log-sums of exponentials, with a latent space of constant size, enabling application with constant time and space complexity per token. We implement our modification, verify that it works in practice, and conclude that it is a promising alternative to conventional attention.\footnote{Source code and instructions for replicating our results are online at \href{https://github.com/glassroom/heinsen_attention}{https://github.com/glassroom/heinsen\_attention}.}
\end{abstract}

\section{Summary}\label{sec:summary}

The conventional attention mechanism of Transformers \cite{DBLP:journals/corr/VaswaniSPUJGKP17} has become the predominant method for capturing sequential dependencies, but its cost is quadratic in sequence length $n$, because it applies a Softmax function over the rows of an $n \times n$ matrix of scaled dot-products. FlashAttention \cite{dao2022flashattention} reduces memory use from quadratic to linear by computing, normalizing, and reducing the matrix in an incremental fashion, {\em i.e.}, without ever storing it in full, but the compute cost remains quadratic.

Numerous alternatives to conventional attention have been proposed to reduce its quadratic cost, including linearized, low-rank, and sparse approximations, mechanisms that slide context windows, and convolutions in the frequency domain. We cannot summarize all proposals fairly in the space of a paragraph. Notable examples include the methods proposed by \citet{DBLP:journals/corr/abs-1904-10509}, \citet{DBLP:journals/corr/abs-2006-04768}, \citet{DBLP:journals/corr/abs-2001-04451}, \citet{DBLP:journals/corr/abs-2006-16236}, \citet{DBLP:journals/corr/abs-2105-14103}, \citet{DBLP:journals/corr/abs-2003-05997}, \citet{DBLP:journals/corr/abs-2105-03824}, and \citet{poli2023hyena}. More recently, generalized state space models that build on previous research \cite{DBLP:journals/corr/abs-1709-04057} \cite{DBLP:journals/corr/abs-2111-00396} have shown promise by incorporating data-driven mechanisms to control the evolution of a fixed-size latent state \cite{peng2023rwkv} \cite{gu2023mamba} \cite{katsch2023gateloop}, but their performance is inferior on certain tasks ({\em e.g.}, recalling arbitrary parts of the input context), motivating the hypothesis that methods with a fixed-size latent space cannot outperform conventional attention \cite{jelassi2024repeat}.

\subsection{Modifying Attention}

We find that a simple modification to conventional attention linearizes it \cite{DBLP:journals/corr/abs-2006-16236} with exponential kernel feature maps, and we show that this modification renders attention expressible as a composition of log-sums of exponentials, with a fixed-size latent space, for sequential application with constant cost per token. We implement our modification, verify that it works, and conclude that it is a promising alternative.

The modification we propose is:

\begin{equation}\label{eq:modified_attention_quadratic}
	\begin{array}{c}
		\modified\Attention(Q, K, V) := \\[0.3em]
		\displaystyle \Softmax\left( \log \frac{\exp(Q) \exp(K)^T}{\exp(c)} \right) V, \\
	\end{array}
\end{equation}

where queries $Q$, keys $K$ and values $V$ have $n_Q \times d_K$, $n_K \times d_K$, and $n_K \times d_V$ elements, respectively, and $c$ is a scalar constant, all in $\mathbb{R}$. We compute all exponentials elementwise.

\subsection{As Log-Sums of Exponentials}\label{ssec:as_log_sums_of_exponentials}

In Section \ref{sec:proof}, we prove that

\begin{equation}
	\thinmuskip=1mu\medmuskip=2mu\thickmuskip=3mu
	\modified\Attention(Q, K, V) = \exp(\log S - \log Z),
\end{equation}

where

\begin{equation}\label{eq:modified_attention_log_S_and_log_Z}
	\thinmuskip=1mu\medmuskip=2mu\thickmuskip=3mu
	\begin{aligned}
		\log S & = \lse_{[d_K]}(Q + \underbrace{\lse_{[n_K]}(K^T + \log V)}_{d_K \times d_V \text{ elements}}) \\[0.5em]
		\log Z & = \lse_{[d_K]}(Q + \underbrace{\lse_{[n_K]}(K^T)}_{d_K \text{ elements}}). \\
	\end{aligned}
\end{equation}

The elementwise sums are over compatible dimensions, broadcasting over all other dimensions, from left to right---{\em e.g.}, before reduction, the broadcasted elementwise sum $K^T + \log V$ has $d_K \times n_K \times d_V$ elements. The functions $\lse_{[d_K]}(\cdot)$ and $\lse_{[n_K]}(\cdot)$ compute log-sums of exponentials over the dimension indexed by $(1, 2, \dots, d_K)$ and $(1, 2, \dots, n_K)$, respectively. If any of $V$'s elements are negative, $\log V$ is complex, and therefore so is $\log S$, but all Softmax mixtures of $V$ remain over $\mathbb{R}$ because they are a composition of operations under which $\mathbb{R}$ is closed \eqref{eq:modified_attention_quadratic}.

\subsection{Autoregressive Case}\label{ssec:autoregressive_case}

For autoregressive attention, in which $n_Q = n_K$ and for each query at step $t$ we compute attention only over $t$ trailing tokens, we note that in \eqref{eq:modified_attention_log_S_and_log_Z}, all sequential dependencies are modeled by the log-sums computed with $\lse_{[n_K]}(\cdot)$, so we can compute autoregressive $\log S$ and $\log Z$ with:

\begin{equation}\label{eq:autoregressive_log_S_and_log_Z}
	\thinmuskip=1mu\medmuskip=2mu\thickmuskip=3mu
	\begin{aligned}
		\log S & = \lse_{[d_K]}(Q + \underbrace{\lcse_{[n_K]}(K^T + \log V)}_{d_K \times n_K \times d_V \text{ elements}}) \\[0.5em]
		\log Z & = \lse_{[d_K]}(Q + \underbrace{\lcse_{[n_K]}(K^T)}_{d_K \times n_K \text{ elements}}), \\
	\end{aligned}
\end{equation}

where the function $\lcse_{[n_K]}(\cdot)$ computes a log-cumulative-sum of exponentials over the dimension indexed by and $(1, 2, \dots, n_K)$.

For sequential application, given a new query $Q_t$ at step $t$, we need only the end-states of the two log-cumulative-sums of exponentials:

\begin{equation}\label{eq:autoregressive_attention_sequential}
	\begin{aligned}
		\log S_t & = \lse_{[d_K]} \big(Q_t + \! \underbrace{\HofS_t}_{d_K \times d_V} \!\! \big) \\[0.5em]
		\log Z_t & = \lse_{[d_K]} \big(Q_t + \underbrace{\HofZ_t}_{d_K} \big), \\
	\end{aligned}
\end{equation}

where hidden states $\HofS_t$ and $\HofZ_t$ are the states of the two log-cumulative-sums at step $t$:

\begin{equation}\label{eq:sequential_attention_hidden_states}
	\begin{aligned}
		\HofS_t & = \log\left(\exp\left(\HofS_{t-1}\right) + \exp(K_t + \log V_t)\right) \\
		\HofZ_t & = \log\left(\exp\left(\HofZ_{t-1}\right) + \exp(K_t)\right), \\
	\end{aligned}
\end{equation}

with zeros as their initial condition:

\begin{equation}\label{eq:sequential_attention_initial_cond}
	\begin{aligned}
		\HofS_0 & = \{0\}^{d_K \times d_V} \\
		\HofZ_0 & = \{0\}^{d_K}. \\
	\end{aligned}
\end{equation}

Together, $\HofS_t$ and $\HofZ_t$ hold the latent, or hidden, state of autoregressive attention's computation at step $t$. They enable us to compute autoregressive attention sequentially with constant time and space complexity per token, $\bigO(1)$.

\subsection{Non-Autoregressive Case}\label{ssec:non_autoregressive_case}

For non-autoregressive attention, in which $n_Q$ may differ from $n_K$ and for each query we compute attention over all tokens in the sequence, we compute $\log S$ and $\log Z$ with \eqref{eq:modified_attention_log_S_and_log_Z}.

For sequential application, in which we add a new token to the input context at step $t$, with key $K_t$ and value $V_t$, we compute $\log S$ and $\log Z$ for all queries from the updated hidden states:

\begin{equation}\label{eq:non_autoregressive_attention_sequential}
	\begin{aligned}
		\log S & = \lse_{[d_K]} \big(Q + \HofS_t \big) \\
		\log Z & = \lse_{[d_K]} \big(Q + \HofZ_t \big), \\
	\end{aligned}
\end{equation}

where $\HofS_t$ and $\HofZ_t$ are the hidden states at step $t$ \eqref{eq:sequential_attention_hidden_states}, with zeros as their initial condition \eqref{eq:sequential_attention_initial_cond}.

\section{Proof}\label{sec:proof}

Given a query $q$ and a key $k$ in $\mathbb{R}^{d_K}$, the logarithm of the dot-product of their exponentials is $\log(\sum(\exp(q)\odot\exp(k))) = \lse(q + k)$, where $\odot$ denotes an elementwise product. Log-sums of exponentials are associative and commutative, making the proof fairly straightforward.

For clarity's sake, we walk step-by-step through a sequence of algebraic manipulations. We start by expanding the Softmax function in \eqref{eq:modified_attention_quadratic} and simplifying the resulting expression. We obtain a form of linear attention \cite{DBLP:journals/corr/abs-2006-16236} with exponential kernel feature maps:

\begin{equation}
	\begin{array}{c}
		\displaystyle \Softmax \left( \!\log \frac{\exp(Q) \exp(K)^T}{\exp(c)} \right) V = \\ \\
		\begin{bmatrix}
			\displaystyle \frac{\exp(Q) \exp(K)^T}{\sum_{[n_K]} \exp(Q) \exp(K)^T}
		\end{bmatrix} V, \\
	\end{array}
\end{equation}

where $\sum_{[n_K]}$ normalizes each row to a probability distribution. The scaling constant $\exp(c)$ disappears because it becomes a common divisor of numerator and denominator expressions. Note that the feature function corresponding to the exponential kernel is infinite dimensional. 

Substitute the dot-products of exponentiated queries and exponentiated keys with equivalent explicit summations over elementwise products:

\begin{equation}
	\begin{bmatrix} \displaystyle
		\frac{\sum_{[d_K]} \exp(Q) \odot \exp(K)^T}{\sum_{[n_K]} \sum_{[d_K]} \exp(Q) \odot \exp(K)^T}
	\end{bmatrix} V,
\end{equation}

where the elementwise product $\odot$ is over compatible dimensions, broadcasting over any other dimensions, from left to right, such that the broadcasted elementwise product $\exp(Q) \odot \exp(K)^T$ has $n_Q \times d_K \times n_K$ elements.\footnote{
	Properly, we are computing a tensor product over the non-compatible dimensions, but we describe it as a combination of elementwise multiplication and broadcasting because those operations will be more familiar to more readers.
}

Express matrix multiplication with $V$ as a summation over broadcasted elementwise products:

\begin{equation}\label{eq:elementwise_factorization}
	\frac{\sum_{[n_K]} \sum_{[d_K]} \exp(Q) \odot \exp(K)^T \odot V}{\sum_{[n_K]} \sum_{[d_K]} \exp(Q) \odot \exp(K)^T}.
\end{equation}

Both $\exp(K)^T$ and $V$ have a dimension indexed by $(1, 2, \dots, n_K)$, but $\exp(Q)$ does not, so we can sum over that dimension before broadcast-multiplying elementwise with $\exp(Q)$:

\begin{equation}
	\frac{\sum_{[d_K]} \exp(Q) \odot \sum_{[n_K]} \exp(K)^T \odot V}{\sum_{[d_K]} \exp(Q) \odot \sum_{[n_K]} \exp(K)^T}.
\end{equation}

Define $S$ and $Z$ as the expressions that compute numerators and denominators, respectively,

\begin{equation}
	\thinmuskip=1mu\medmuskip=2mu\thickmuskip=3mu
	\begin{aligned}
		S & := \textstyle \sum_{[d_K]} \exp(Q) \odot \sum_{[n_K]} \exp(K)^T \odot V \\[0.5em]
		Z & := \textstyle \sum_{[d_K]} \exp(Q) \odot \sum_{[n_K]} \exp(K)^T, \\
	\end{aligned}
\end{equation}

and take their logarithms. We obtain:

\begin{equation}
	\thinmuskip=1mu\medmuskip=2mu\thickmuskip=3mu
	\begin{aligned}
		\log S & = \lse_{[d_K]}(Q + \lse_{[n_K]}(K^T + \log V)) \\
		\log Z & = \lse_{[d_K]}(Q + \lse_{[n_K]}(K^T)), \\
	\end{aligned}
\end{equation}

which is the same as \eqref{eq:modified_attention_log_S_and_log_Z}.

\section{Implementation}

As proof of concept, we implement our attention mechanism for both autoregressive applications ({\em e.g.}, generative language modeling) and non-autoregressive applications ({\em e.g.}, masked language modeling). For simplicity and expediency, we limit our implementation in two significant ways: First, we restrict $V$ to elements $\ge 0$ to avoid dealing with complex floating-point numbers, which incur greater overhead and are more cumbersome to manipulate than real floating-point numbers with existing software infrastructure. Second, when computing autoregressive attention over $n_K$ tokens, we first compute all $n_K$ hidden states with a parallel scan, and then reduce them, which is space-inefficient but easier to implement with existing software infrastructure.\footnote{
	It is possible to compute autoregressive attention over $n_K$ tokens in parallel more space-efficiently, by evaluating
	$$\log S = \lse_{[d_K]}( Q + \lcse_{[n_K]}(K^T + \log V))$$
	without simultaneously storing all $d_K \times n_K \times d_V$ intermediate values of the broadcasted sum $K^T + \log V$ in memory.
}

We apply our implementation in a small generative language model (125M parameters, 50257 token ids, 768 embedding features). For numerical stability, in each layer we compute $\log V$ over $\mathbb{R}$ directly, with a dense feed-forward transformation of token states, implicitly defining $V$ as $\log V$'s exponential but never actually computing it. To remain in $\mathbb{R}$, we use the logarithm of attention as input to subsequent transformations in the layer, {\em i.e.}, the input to subsequent transformations is $\log S - \log Z$ instead of $\exp(\log S - \log Z)$. Please see our published code for all model details. We train the model on 300B tokens from The Pile \cite{pile} with a conventional sequence length of 1024 tokens, and obtain a cross-entropy loss of 2.47, competitive with state-of-the-art generative language models of similar size.

\section{Conclusions}

By all indications, our attention mechanism is a promising alternative to the conventional one, but the evidence we have so far is too scant to be conclusive. An adequate comparison requires addressing our implementation's temporary limitations and evaluating models with one to several orders of magnitude more parameters on a diverse set of benchmarks and downstream tasks.

\cleardoublepage

\bibliography{main}
\bibliographystyle{main}

\end{document}